\documentclass[bst/sn-nature]{sn-jnl}

\usepackage{graphicx}%
\usepackage{multirow}%
\usepackage{amsmath,amssymb,amsfonts}%
\usepackage{amsthm}%
\usepackage{mathrsfs}%
\usepackage[title]{appendix}%
\usepackage{xcolor}%
\usepackage{textcomp}%
\usepackage{manyfoot}%
\usepackage{booktabs}%
\usepackage{algorithm}%
\usepackage{algorithmicx}%
\usepackage{algpseudocode}%
\usepackage{listings}%
\usepackage[numbers]{natbib}
\usepackage{comment}

\theoremstyle{thmstyleone}%
\theoremstyle{thmstyletwo}%

\theoremstyle{thmstylethree}%

\begin{document}

\title[Article Title]{FengWu-GHR: Learning the Kilometer-scale Medium-range Global Weather Forecasting}


\author[1]{\fnm{Tao} \sur{Han}}

\author[2]{\fnm{Song} \sur{Guo}}
\author[3,1]{\fnm{Fenghua} \sur{Ling}}
\author[1]{\fnm{Kang} \sur{Chen}}
\author[1]{\fnm{Junchao} \sur{Gong}}
\author[3]{\fnm{Jingjia} \sur{Luo}}
\author[4]{\fnm{Junxia} \sur{Gu}}
\author[5]{\fnm{Kan} \sur{Dai}}
\author[1]{\fnm{Wanli} \sur{Ouyang}}
\author*[1]{\fnm{Lei} \sur{Bai}}\email{baisanshi@gmail.com}


\affil[1]
{\orgname{Shanghai AI Laboratory}}
\affil[2]
{\orgname{Hong Kong University of Science and Technology}}
\affil[3]
{\orgname{Nanjing University of Information Science and Technology}}
\affil[4]
{\orgname{National Meteorological Information Center of CMA}}
\affil[5]
{\orgname{National Meteorological Center of CMA}}


\abstract{ Kilometer-scale modeling of global atmosphere dynamics enables fine-grained weather forecasting and decreases the risk of disastrous weather and climate activity. Therefore, building a kilometer-scale global forecast model is a persistent pursuit in the meteorology domain. Active international efforts have been made in past decades to improve the spatial resolution of numerical weather models. Nonetheless, developing the higher resolution numerical model remains a long-standing challenge due to the substantial consumption of computational resources. Recent advances in data-driven global weather forecasting models utilize reanalysis data for model training and have demonstrated comparable or even higher forecasting skills than numerical models. However, they are all limited by the resolution of reanalysis data and incapable of generating higher-resolution forecasts. This work presents FengWu-GHR, the first data-driven global weather forecasting model running at the 0.09$^{\circ}$ horizontal resolution. FengWu-GHR introduces a novel approach that opens the door for operating ML-based high-resolution forecasts by inheriting prior knowledge from a pretrained low-resolution model. The hindcast of weather prediction in 2022 indicates that FengWu-GHR is superior to the IFS-HRES. Furthermore, evaluations on station observations and case studies of extreme events support the competitive operational forecasting skill of FengWu-GHR at the high resolution.}




\maketitle

\section{Introduction}\label{sec1}
Numerical Weather Prediction (NWP) is the most crucial pillar in meteoroloy, where High-Resolution (HR) operational weather forecasting is regarded as one aspect of improvements in skills scores~\cite{rodwell2010new}. Intuitionally, HR forecast can provide more spatially accurate weather service for daily human life and social production, thereby enabling the guidance function of the weather forecast in a more fine-grained way. In addition, HR could be beneficial for weather and climate research. By utilizing high-resolution models, researchers can more accurately capture finer-scale atmospheric processes and phenomena. This enables more precise predictions and simulations of severe weather events such as heat waves~\cite{iles2020benefits}, floods~\cite{younis2008benefit,lagasio2019synergistic,van2015resolution}, drought~\cite{van2015resolution_erl}, and storms~\cite{willison2015north}. The improved resolution allows for a better representation of the intricate dynamics and interactions that drive these events, ultimately enhancing our ability to forecast their intensity and impacts. Growing evidence demonstrates that HR improves the process chain, thus resulting in enhanced global weather forecasting~\cite{roberts2018benefits}.

In the past few years, machine learning methods(ML)~\cite{bi2022pangu,lam2022graphcast,chen2023fengwu,kochkov2023neural,kurth2023fourcastnet} show the potential to surpass the state-of-the-art physical-based NWP model~\cite{charney1950numerical,lynch2008origins} in terms of forecast accuracy and operational efficiency. Yet now, there remains a significant gap in spatial resolution between artificial intelligence (AI) models and the current operating version (Cycle 48r1 \cite{81380}) of the Integrated Forecasting System (IFS). The coarse resolution ($0.25^{\circ}\times0.25^{\circ}$ or below) is hard to simulate the small-scale atmospheric processes, such as convective storms and orographic rainfall~\cite{sillmann2017understanding}. Although some previous studies try to level up the resolution with downscale methods~\cite{luo2024diffusion,gao2021downscaling}, they generally take it as a post-processing and could not solve the fundamental problem of smaller-scale meteorological phenomena. Therefore, developing a high-resolution global weather forecast model is in great need. 

Two primary challenges prevent ML-based global weather forecast models from upgrading to a high spatial resolution model. The first challenge is the lack of large-scale datasets at the global scale with high resolution, which fundamentally limits the development of HR data-driven weather forecast models. For example, the European Center for Medium-Range Weather Forecasts (ECMWF) only provides HR products from 2016\footnote{https://www.ecmwf.int/en/newsletter/147/meteorology/new-model-cycle-brings-higher-resolution}, leading to much less available data compared with the ERA5 reanalysis. Second, although the inference time of ML models is significantly faster than physical-based models~\cite{ben2023rise}, they still encounter expensive computational costs during the training phase. For example, the training of GraphCast~\cite{lam2022graphcast} takes 21 days on 32 Cloud TPU v4, with a spatial resolution of $721\times1440$. In general, the computational complexity is $O(N^{2})$ when increasing the spatial grid resolution to $N\times$. Hence, suppose that we have access to the copious HR data, the potential investment required to train HR models would remain substantial, notwithstanding leveraging cutting-edge computing facilities~\cite{kochkov2023neural}.

Despite the formidable challenges faced in such a period, the presentation of FengWu-GHR marks a successful endeavor in spearheading the transition of large-scale weather forecasting models into the high-resolution era. Specifically, there are three notable highlights of FengWu-GHR in comparison to the existing ML-based global weather forecast models. \textbf{Highest resolution}: FengWu-GHR is the first AI weather forecasting model operating at a high-resolution grid 
($2001\times4000$ in latitude and longitude, around 9km~\cite{rasp2023weatherbench}), aligning closely with the operational version of IFS-HRES. This advancement effectively enhances the grid density by approximately eightfold compared with the existing advanced AI models. \textbf{Largest learnable parameters}: FengWu-GHR currently stands as the most parameter-rich forecasting model in atmospherical science, boasting a total count of trainable parameters exceeding 4 billion. \textbf{General framework for downscaling}: The insights behind FengWu-GHR have strong transferability, enabling the advancement of many ML-based weather forecast models from low resolution to high resolution. Particularly, it allows the current models to be upgraded with minimal cost while preserving excellent performance.

In summary, the novel model design and optimization strategy of Fengwu-GHR not only improve the resolution but also further boost the accuracy of medium-range weather forecasting in the following aspects:
1) \textbf{Fine-grained global forecast}: In comparison with both physical and data-driven global weather forecast models, we demonstrate that FengWu-GHR has better consistency with the real-time analysis data, which means FengWu-GHR can generate more accurate and detailed weather prediction than other models. 2) \textbf{Accurate station prediction}: We assess the predictive ability of the ML-based 
 weather forecast model on large-scale station observations according to the real operational process, revealing that FengWu-GHR achieves the smallest error and higher resolution is more suitable for station-level weather forecasting. 3) \textbf{Long lead-time stability}: FengWu-GHR reduces the bias drifts problem of medium-range weather forecast that exist in many ML-based global weather forecast models~\cite{chen2023fengwu,lam2022graphcast,bi2022pangu}. 4) \textbf{Reliable extreme forecast}: As for extreme weather forecasts, we also observe that FengWu-GHR can perceive the occurrence of extreme events such as heat waves and winter storms earlier and with greater accuracy.

\section{FengWu-GHR}
\begin{figure}[ht!]
  \centering      \includegraphics[width=0.99\textwidth]{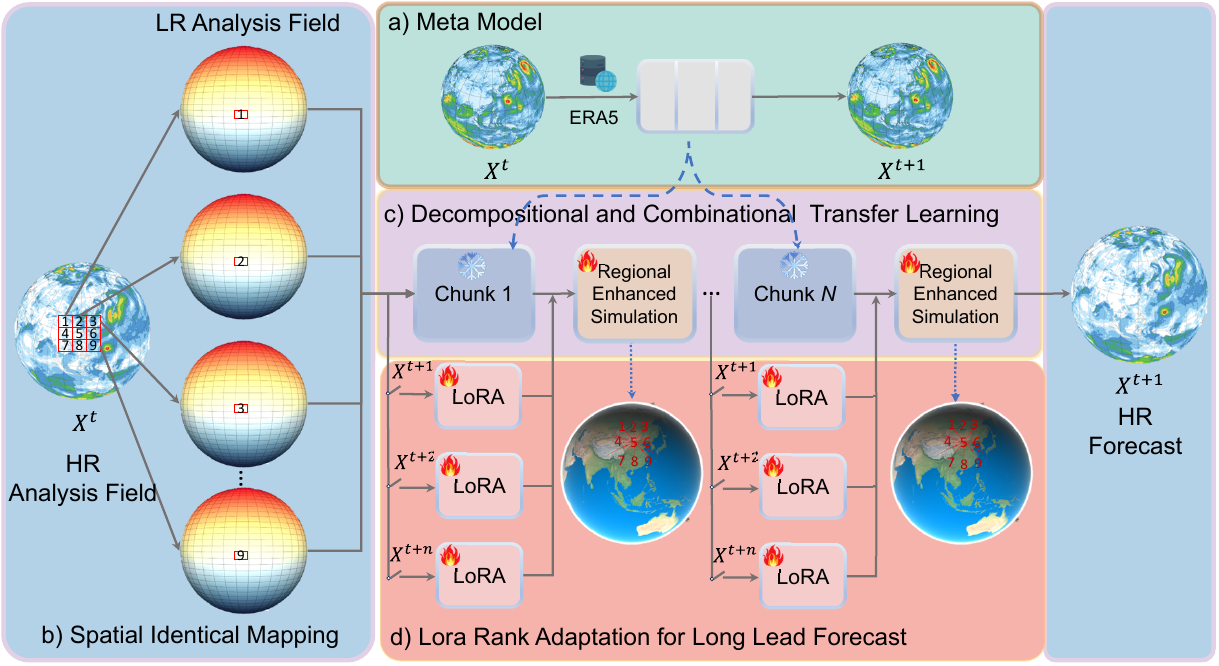}
    \caption{\textbf{Structutr of FengWu-GHR}. a) The first stage is to learn the physical laws from the long-term historical reanalysis data at a low resolution. b) An extrapolation method that enables pretrained LR models to operate on high-resolution analysis fields. c) A transfer learning algorithm that can complement the pretrained model to capture the small-scale weather phenomena. d) A low-rank adaptation that is implemented on the parameter level for each step to improve the forecast skill during long roll-out.}
    \label{fig:framework}
\end{figure}

This section elucidates the methodology for learning kilometer-scale medium-range global weather forecasting, encompassing model architecture and training procedures. Notably, as a groundbreaking endeavor in high-resolution forecasting, FengWu-GHR simulates the evolution of atmospheric variables akin to FengWu~\cite{chen2023fengwu}, including five pressure level variables (each variable has 13 vertical levels): geopotential ($z$), specific humidity ($q$), zonal component of wind ($u$), meridional component of wind ($v$), and air temperature ($t$), and four surface variables: 2-meter temperature (t2m), 10-meter u wind component (u10), 10-meter v wind component (v10), and mean sea level pressure (msl).   

\subsection{Meta Model}
Current ML-based global weather forecast models are mainly classified into two prominent families: the Transformer-based~\cite{kurth2023fourcastnet, chen2023fengwu, bi2022pangu, fuxi}, and Graph Neural Network categories~\cite{keisler2022forecasting,lam2022graphcast, price2023gencast}. Both of these structures can provide powerful spatiotemporal representation capabilities~\cite{vaswani2017attention, pfaff2020learning,sanchez2020learning}. Conventionally, GNN-based networks are extensively employed to comprehend intricate physical dynamics of fluids and other graph systems~\cite{lam2022graphcast, pfaff2020learning}, while transformer-based networks have proliferated across various domains, such as computer vision~\cite{han2022survey}, natural language processing~\cite{vaswani2017attention}, and even multimodal learning~\cite{liu2023visual}, resembling the rapid growth of mushrooms after rainfall. A notable advantage of the transformer-based network lies in its remarkable versatility and scalability, as exemplified by the astounding achievements of large language models~\cite{brown2020language,touvron2023llama}. 

The core component of FengWu-GHR is the meta model, designed with principles of simplicity, expansibility, and scalability. Thereby, the meta model is primarily based on the attention-based vision transformer~\cite{dosovitskiy2020image}, and it consists of three key components: a 2-D patch embedding layer, stacked transformer blocks, and a deconvolution layer. As depicted in Figure~\ref{fig:framework}, the input to the meta model is an initial weather state, denoted as $X^{t}\in{\mathbb{R}^{C\times H \times W}}$, where $C\times H \times W$ represents the stacked weather state with multiple levels of upper air and surface variables, in which vertical latitude and horizontal longitude are divided into $H$ and $W$ grids for each variable.

The patch embedding layer takes in the multi-dimensional weather state and encodes it into a sequence representation denoted as $S\in \mathbb{R}^{N \times D}$, where D is the number of predefined feature dimension and the value of $N$ is determined by applying a convolution stride of $(P, P)$ in the embedding layer, resulting in $N=\frac{H}{P} \times \frac{W}{P}$. The subsequent transformer blocks comprise standard self-attention and feed-forward modules~\cite{vaswani2017attention,dosovitskiy2020image}. The meta model consists of $M$ blocks to further process the sequence, enabling information interaction and feature refinement. Ultimately, the sequence representation is reshaped back into a 2-D spatial representation of dimensions $C \times \frac{H}{P} \times \frac{W}{P}$, and finally restored to its original shape $C\times H \times W$ using a deconvolution upsampling layer. This pipeline facilitates the generation of a one-step forecast for the future weather state denoted as $X^{t+1}$.

Compared with the original ViT~\cite{dosovitskiy2020image}, the proposed meta model incorporates two significant modifications aiming at reducing computational costs and enhancing forecast accuracy. Firstly, to mitigate excessive memory allocation during the self-attention operation, we employ a method that computes cross-interactions between local-window attention and global-window attention. Specifically, within every $F$ transformer blocks, $F-1$ blocks perform local-window attention. Empirical experiments demonstrate that this approach does not result in noticeable accuracy reduction while significantly reducing memory usage and expediting the training process. Secondly, to incorporate prior knowledge, the window attention is progressively applied to specific regions, including the square, zonal rectangle, and meridional rectangle. This strategic implementation allows the model to incorporate relevant climate zone prior during the forecasting process.

\subsection{Spatial Identical Mapping Extrapolate}
In FengWu-GHR, the meta model serves as a crucial component that is pretrained on a relatively lower grid resolution (e.g., $721 \times 1440 $) using approximately 42 years (1979-2021) of ECMWF's ERA5 reanalysis dataset~\cite{hersbach2020era5}. Our objective is to extend this pretrained model's capability to perform cross-resolution forecasting. However, the existing extrapolation methods~\cite{su2024roformer,press2021train,sun2022length} for AI models predominantly focus on NLP tasks, rendering them unsuitable for atmospheric data due to the fundamental difference in embedding operations. While NLP embedding involves querying sequences from a vocabulary~\cite{mikolov2013efficient}, atmospheric data lacks such a vocabulary for embedding purposes. Consequently, the intuitive approach of interpolating the initial field to match the pretrained model's input and subsequently upsampling the forecast to a higher resolution yields unsatisfactory results.

The extrapolation of transformer-based atmospheric prediction models faces two major challenges. Firstly, the representative area for each token remains fixed once the embedding layer's convolutional kernels are trained. For instance, a token embedded by an $8\times8$ 2-D kernel at a grid resolution of $0.25^{\circ}\times0.25^{\circ}$ approximately represents a geographical area of $240 \text{km} \times 240 \text{km}$. However, when the input data is at a grid resolution of $0.09^{\circ}\times0.09^{\circ}$, the token represents an area of only $80 \text{km} \times 80 \text{km}$, leading to inconsistent mappings and a significant degradation in performance. Furthermore, the sequence length increases to $9 \times N$, posing tremendous computational challenges due to the quadratic complexity $O((9N)^2)$ of global self-attention modules. Considering these challenges, we propose the Spatial Identical Mapping Extrapolate (SIME) method as an innovative approach to ensure reasonable skill scores when applying the meta model to high-resolution initial fields.

As depicted in Figure~\ref{fig:framework}, the SIME method, when applied to a high-resolution (HR) initial field with a spatial resolution of $X_{h} \in \mathbb{R}^{H_{h} \times W_{h}}$, operates under the fundamental thought of decomposing it into a batch of low-resolution (LR) initial fields. These LR initial fields, denoted as $X_{l} \in \mathbb{R}^{B \times H \times W}$, are at a scale that the meta model has been trained on. Consequently, the problem of extrapolating a large-scale field is transformed into inferring multiple LR initial fields. To ensure that each sample in $X_{l}$ has a similar spatial representation as the training data, we propose the spatial identical mapping strategy to disassemble $X_{h}$. For instance, assume that $H_{h} \times W_{h} = 9 \times H \times W$, each grid in $X_{l}$ can be viewed as an ensemble of a $3\times3$ patch in $X_{h}$. Within this patch, each grid has the shortest distance to the corresponding grid in $X_{l}$ compared to other grids in $X_{h}$. As a result, each grid in the patch can be mapped to the same spatial position in $X_{l}$. Similarly, $X_{h}$ is divided into numerous patches with a stride of $3 \times 3$, and each patch can be mapped to a corresponding grid in $X_{l}$. By assembling the grids from different patches that occupy the same positions, we obtain several large-scale initial fields. Each of these fields maintains an identical geographical spatial resolution as the training data, ensuring a plausible extrapolated result. Notably, the SIME method achieves this with a reduced computational complexity of $O(9N^2)$, resulting in a ninefold decrease in computational requirements.

\subsection{Decompositional and Combinational  Transfer Learning}
Despite the effectiveness of SIME in enabling the meta model to utilize higher-resolution initial fields and produce acceptable results, it can not improve skill scores, as it overlooks small-scale weather phenomena present in the HR initial field. This limitation can be attributed to two primary factors. Firstly, the meta model is solely trained on LR reanalysis data, resulting in a domain gap when performing cross-domain inference with the operational analysis field. Consequently, this cross-domain inference leads to a reduction in performance. Secondly, SIME treats those grids within a patch independently, disregarding the interdependencies and convective phenomena within the patch. To address these problems, one feasible solution is to incorporate transfer learning into the meta model using HR operational analysis data. By leveraging this transfer learning approach, the meta model can effectively capture the small-scale convective phenomena and bridge the domain gap, thereby enhancing its skill scores for high-resolution forecasting.

As illustrated in Figure~\ref{fig:framework}, we modify the meta model to incorporate Regional Enhanced Simulation (RES) modules, which are designed to capture small-scale weather phenomena by conducting Decompositional and Combinational Transfer Learning (DCTL). Typically, a RES module that performs a local attention mechanism is inserted after a few transformer blocks. When the decomposed input sequence $S \in \mathbb{R}^{B\times N \times D}$ enters a RES module, all tokens are rearranged to their original spatial positions. In other words, the sequence's shape is restored to $\sqrt{B}\frac{H}{P} \times \sqrt{B}\frac{W}{P} \times {D}$. Subsequently, these tokens, distributed in a 2-D spatial layout, are divided into different regions using a window partition operation. Each region then undergoes a multi-head self-attention~\cite{vaswani2017attention} operation to enhance the learning of small-scale weather patterns. Afterward, the output of the RES module is transformed back to the shape of $S \in \mathbb{R}^{B\times N \times D}$ using a window reverse operation, ensuring compatibility with the subsequent processing steps. 

Due to the HR analysis data being only accessible since 2016, to avoid overfitting, we only embed several RES modules into the meta model. Nevertheless, even with this constraint, we observe substantial performance improvements after performing DCTL. This transfer learning approach empowers the meta model with the capability to effectively capture and represent intricate weather patterns and fine-grained weather information.

\subsection{Low-rank Adaptation for Long-lead Forecasting}
The meta model of FengWu-GHR runs with a lead time of six hours, necessitating the adoption of an autoregressive strategy to generate long-lead forecasts. To mitigate error accumulation during the roll-out process, previous studies~\cite{lam2022graphcast,kurth2023fourcastnet,chen2023fengwu, fuxi} have employed various methods to fine-tune pretrained models. For instance, GraphCast~\cite{lam2022graphcast} progressively fine-tunes each forecasting step for a proportional number of iterations. FengWu introduces a reply buffer~\cite{chen2023fengwu} to store the model predictions during training and reuse them for training, which enables the optimization of long-lead forecasting implicitly. However, these methods primarily focus on fine-tuning a shared model across multiple steps, without considering the potential conflicts in different steps. In this context, FengWu-GHR proposes the utilization of the Low-Rank Adaptation (LoRA)~\cite{hu2021lora} method to individually fine-tune each step based on personalized parameters. By employing LoRA, FengWu-GHR addresses the limitations of previous methods and ensures that each step is fine-tuned while preserving the integrity of the previously trained parameters.

We here denote a group of pretrained parameters in the meta-model as the matrix $W_{i}^{0}$. For step $t$, the fine-tuning process only updates its low-rank decomposition, represented as $\Delta W_{i}^{t}=B_{i}^{t}A_{i}^{t}$, where $B_{i}^{t} \in \mathbb{R}^{D \times r}$, $A_{i}^{t} \in \mathbb{R}^{r \times K}$, and $r \ll \text{min}(D, K)$. Consequently, upon completion of the fine-tuning process, the final parameters for step $t$ are updated from $W_{i}^{0}$ to $W_{i}^{0}+\Delta W_{i}^{t}$. Throughout the fine-tuning process, the meta model remains frozen, and the pretrained parameters are fixed. The trainable parameters $A_{i}^{t}$ and $B_{i}^{t}$ are combined with the pretrained parameters when rolling out to each step. This transformation modifies the original feed-forward process $h_{i}^{t}=W_{i}^{0} x$ to:

\begin{equation}
h_{i}^{t}=(W_{i}^{0}+ \beta \Delta W_{i}^{t}) x,
\end{equation}
where $\beta=\frac{\alpha}{r}$ is a scale factor. During the fine-tuning process, $A_{i}^{t}$ is initialized using random Gaussian values, while $B_{i}^{t}$ is initialized with zeros. This initialization ensures that each step undergoes fine-tuning starting from the state of the meta model. By incorporating the Low-Rank Adaptation (LoRA) technique, we can independently customize a set of parameters for each forecasting step. This approach allows for a more precise correction of biases that may arise during the long roll-out process.

\section{Result}\label{sec4}

\subsection{Dataset}\label{subsec2}

\textbf{ERA5}~\cite{hersbach2020era5} is a global atmospheric reanalysis dataset with $0.25^{\circ}$ resolution provided by ECMWF. ERA5 comprises several fundamental variables with 37 vertical levels (\emph{e.g.}, wind, temperature, humidity, and ugeopotential \emph{etc.}) and plenty of surface variables (\emph{e.g.}, temperature at 2-meter (t2m), wind speed at 10-meter (u10 and v10), mean sea level pressure (msl), precipitation \emph{etc.}), which makes it suitable for a wide range of applications, including climate research, weather forecast, and environmental monitoring. It is considered reliable by the field of data-driven approaches, which frequently utilize ERA5 as a training database.

\noindent \textbf{Real-time Analysis Data} is regarded as the initial conditions to operate the IFS-HRES forecast, which is produced every 6 hours using 4D-Var system. A discussion in WeatherBench2~\cite{rasp2023weatherbench} has pointed out that it is unfair to evaluate the IFS-HRES forecast against ERA5 as the non-zero error occurred at time step $t=0$. Hence, ERA5 is only used as the pre-train material in this study, and the final evaluation is performed on the official operational analysis data. The official analysis is stored in the ECMWF archive at a high resolution ( $\sim$9km) and the ECMWF's internal evaluation uses this analysis data. The analysis data is stored in spectral or Gaussian reduced format and interpolated to the latitude and longitude grid with internal interpolation software. For the analysis data, we use the data from 2016, a time point when the IFS was upgraded to 9km, to 2021 for training, and 2022 for evaluation.

\subsection{Verificatoin Metrics}\label{subsec4}
\label{sec:metric}
There is a set of standard evaluation processes used to evaluate weather forecast models in ECMWF and WeatherBench~\cite{ben2023rise,rasp2023weatherbench}, among which scores composed of weighted Root Mean Square Error (RMSE) and weighted Anomaly Correlation Coefficient (ACC) have been widely used by ML-based prediction models~\cite{bi2022pangu,chen2023fengwu,lam2022graphcast}.
In addition, the averaged difference between forecasts and observations is used to measure the statistical bias of the model~\cite{ben2023rise,rasp2023weatherbench}, while the activity is defined as the standard deviation of forecast/observation anomalies to evaluate the smoothing problem in model forecasts~\cite{ben2023rise}.

Note that all metrics are computed using an area-weighting over grid points due to the non-equal area distribution from the equator towards the north and south poles. The latitude weights $\alpha(i)$ are computed as,
\begin{equation}
\label{eq:MSE}
\alpha(w) = \frac{W\operatorname{cos}(\theta_{i})}{\sum_{j}^W \operatorname{cos}(\theta_{j})},
\end{equation}
where $\theta_{i}$ 
represents the degree value corresponding to the $i_{th}$ longitude, ranging from -90° to +90°. $W$ is the number of latitudes in a given resolution.  

\textbf{RMSE} is a metric commonly used in meteorology to directly assess the accuracy of the forecasts. Given the prediction result $\hat{x}_{c, w, h}^{i+\tau}$ and its target (ground truth) $x_{c, w, h}^{i+\tau}$, the RMSE is defined as follows:

\begin{equation}
\label{eq:MSE}
\operatorname{RMSE}(c,\tau ) =\frac{1}{T} \sum_{t}^{T} \sqrt{\frac{1}{WH}\sum_{w,h} \alpha(w)(x_{c,w,h}^{i+\tau} - \hat{x}_{c,w,h}^{i+\tau})^{2}},
\end{equation}
where $c$ denotes the different variables at the surface level or pressure level. $\tau$ is the lead time of forecasts. $T$ is the total number of valid test time slots. 

\textbf{ACC} evaluates the performance of gloabl weather forecast models by comparing their predictions of anomalies (departures from the long-term averaged climatology) to observed anomalies,

\begin{equation}
\label{eq:ACC}
\operatorname{ACC}(c,\tau ) =\frac{1}{T} \sum_{i=1}^{T} \frac{\sum_{w,h} \alpha(w) (x_{c,w,h}^{i+\tau} - C_{c,w,h}^{i+\tau}) (\hat{x}_{c,w,h}^{i+\tau} - C_{c,w,h}^{i+\tau})}{
\sqrt{\sum_{w,h} \alpha(w) (x_{c,w,h}^{i+\tau} - C_{c,w,h}^{i+\tau})^{2} \sum_{w,h}  \alpha(w) (\hat{x}_{c,w,h}^{i+\tau} - C_{c,w,h}^{i+\tau})^{2}}
},
\end{equation}
where $C_{c,w,h}^{i+\tau}$ is the climatology.
Here we follow GraphCast~\cite{lam2022graphcast} and FengWu~\cite{chen2023fengwu} and compute the climatology $C$ as a function of the day of year by taking the mean of operational analysis data at the $0.09^{\circ}$ latitude/longitude grids from 2016 to 2021 for each grid point.

\textbf{Bias} is the mean error computed for each location and then averaged on the point-wise error,
\begin{equation}
\label{eq:MSE}
\operatorname{Bias}(c,\tau ) =\frac{1}{T} \sum_{t}^{T} \sqrt{\frac{1}{WH}\sum_{w,h} \alpha(w)(x_{c,w,h}^{i+\tau} - \hat{x}_{c,w,h}^{i+\tau})}.
\end{equation}

Normally, the bias indicates the stability of a forecasting system when continuously running the model for a long lead time. However, it can not reflect the performance of extreme events as positive and negative biases can cancel each other out, resulting in changes primarily observed in the mean state at different forecast lead times.

\textbf{Activity} is a metric to measure the smoothness of the forecasts, the lower the forecast activity the smoother the forecast. 

\begin{equation}
\small
\label{eq:MSE}
\operatorname{Act.}(c,\tau ) =\frac{1}{T} \sum_{t}^{T} \sqrt{\frac{1}{WH}\sum_{w,h} \alpha(w)\left[(x_{c,w,h}^{i+\tau} - \hat{x}_{c,w,h}^{i+\tau}) - \frac{1}{WH} \sum_{w,h} \alpha(w)(x_{c,w,h}^{i+\tau} - \hat{x}_{c,w,h}^{i+\tau})\right]},
\end{equation}

\subsection{Global weather forecast verification on operational analysis}

\begin{figure}[htbp]
  \centering      
   \includegraphics[width=0.92\textwidth]{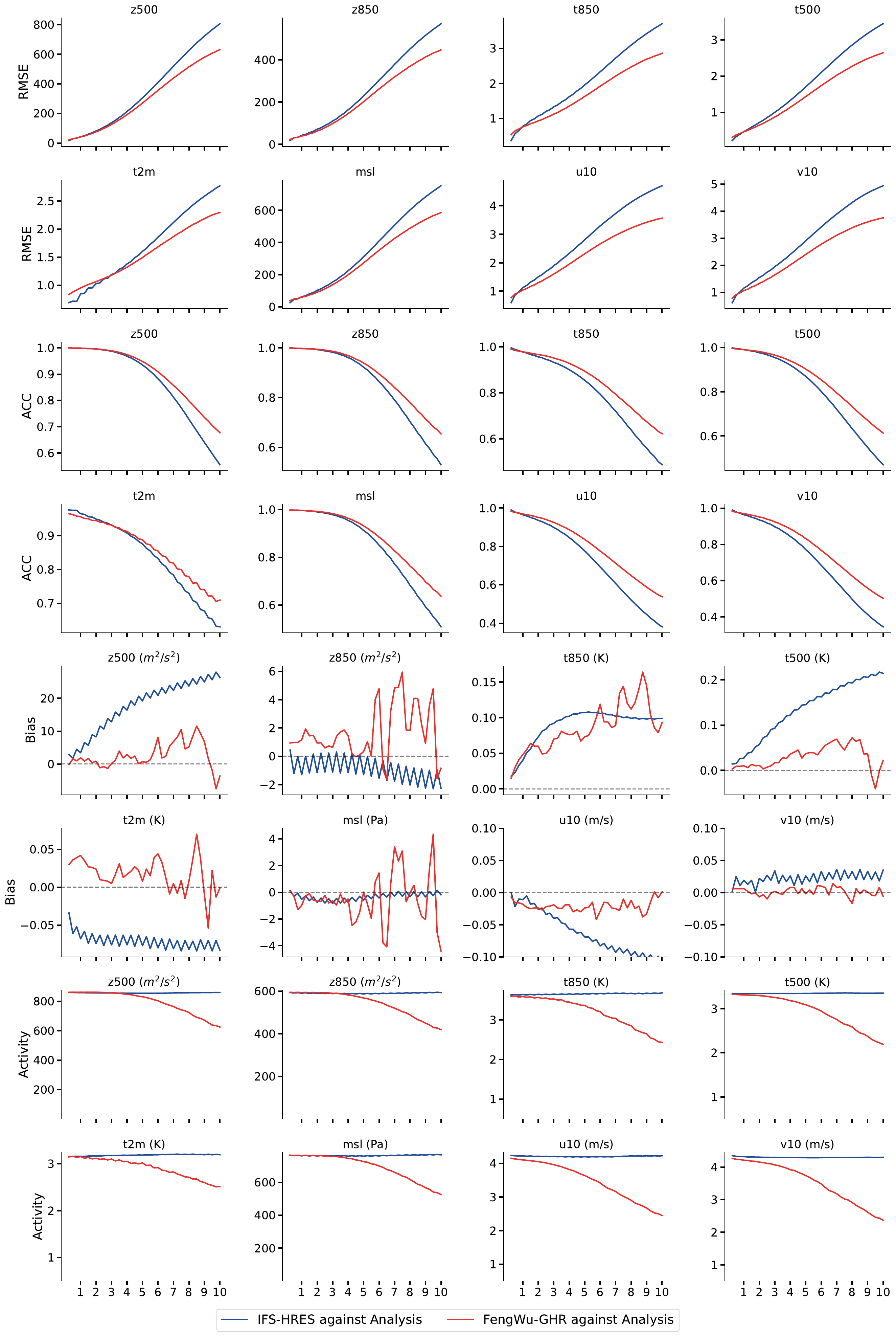}
    \caption{\textbf{
     The predictive skills of FengWu-GHR and IFS-HRES in weather forecast during 2022}. RMSE: the lower the better. ACC: the higher the better, Bias: the closer to zero the better. Activity: the lower the forecast activity the smoother the forecast. The forecast evaluation metric of FengWu-GHR and IFS-HRES are both computed against the operational analysis data at the grid of $0.09^\circ \times 0.09^\circ$. The x-axis in each sub-figure represents lead time, at a 6-hour interval over a 10-day forecast.
     }
    \label{fig:result_RMSE_ACC}
\end{figure}

 FengWu-GHR and the most advanced version of IFS-HRES ~\footnote{https: //confluenceecmwf.int/display/FCST/Changes+to+the+forecasting+system} are compared using four metrics introduced in Section~\ref{sec:metric}. The evaluation aimed to ensure a fair comparison by evaluating both models using operational analysis data, which is closer to real-time forecasting. As IFS-HRES is gradually updated, the closer the data is to the present, the more accurate it is considered, thereby the evaluation specifically focused on the year 2022.
Because the volume of IFS-HRES forecast data is very large, the comparison is performed on a subset of variables in different pressure levels, specifically 320 variables (8 variables $\times$ 40 steps) for each metric. By analyzing the results of RMSE nd ACC for 2022, it is found that FengWu-GHR outperforms IFS-HRES in $91.3\%$ of the 320 target variables.

Figure~\ref{fig:result_RMSE_ACC} highlights the superiority of FengWu-GHR (red lines) in weather forecast compared to IFS-HRES (blue lines) in terms of the RMSE skill and ACC skill. 
Take the z500 (Geopotential at 500 hPa), an important atmospherical variable associated with many meteorological factors~\cite{xoplaki2000connection,turkes2002persistence}, the RMSE shows FengWu-GHR is overall lower than IFS-HRES. Specifically, the RMSE skill score has been improved by around $11.5\% - 21.7\%$ from 5 to 10 lead days. Other variables also demonstrate the same trends of skill improvements, e.g., the temperature's RMSE at 850 hPa drops from 3.76 to 2.86 at 10 days lead, showing a substantial ascent for temperature forecast.         

It is widely recognized that ML-based weather forecast models often experience significant bias drift with lead-time~\cite{rasp2023weatherbench, ben2023rise}.  Figure~\ref{fig:result_RMSE_ACC} shows some variables' average bias over one year. For this metric, an increasing bias indicates that the weather forecast model moves away from its initial conditions and toward its long-term statistical mean. Typically this new statistical mean has some unrealistic features. Notably, FengWu-GHR has less bias in some variables such as u10, v10, t2m, z500, and t500, while the strong bias drift for z500 is found for Pangu-weather~\cite{ben2023rise}. These advancements signify that ML-based models can also achieve comparable stability of long-term weather forecasts with good optimization and model design.   

As mentioned in the research of ECMWF~\cite{ben2023rise}, it is important to check the level of activity of the different forecasts when comparing RMSE from different models. 
Compared to IFS-HRES, the majority of variables in FengWu-GHR exhibit decreased activity as lead time increases (Figure~\ref{fig:result_RMSE_ACC}), indicating a reduction in the amplitude of weather variations over long-term prediction periods. This problem could be attributed to the paradigm of the fine-tune process.

\subsection{Overall forecast evaluation on station observation}

\begin{figure}[htbp]
  \centering      
  \includegraphics[width=0.99\textwidth]{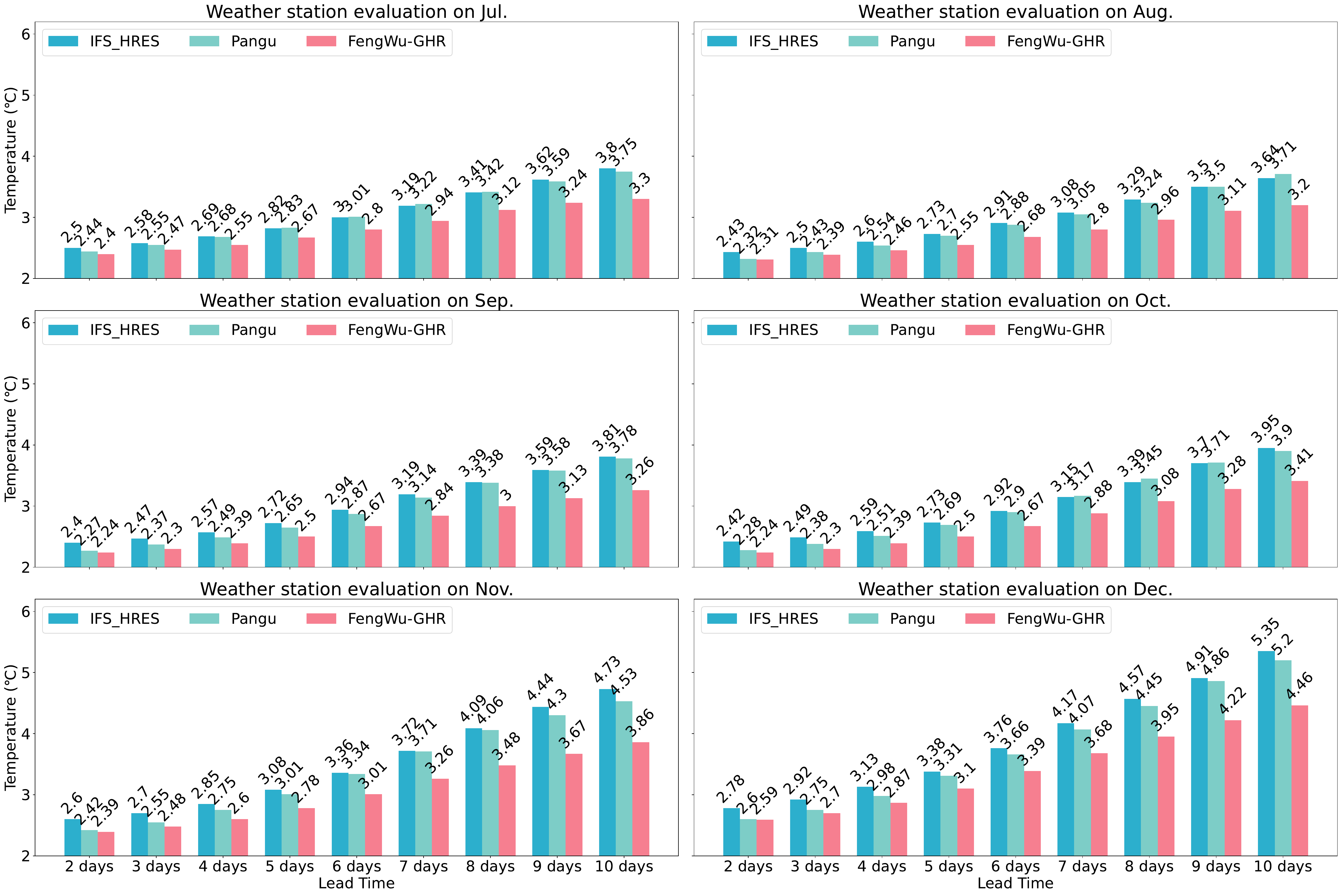}
    \caption{Comparison of RMSE for surface temperature prediction relative to station observations during 2022. The plot displays the performance of IFS-HRES (blue), Pangu-weather (cyan), and FwngWu-GHR (red). 
    }
    \label{fig:observatio_tmp}
\end{figure}

In the operational scenario of weather forecasting, station evaluation of forecast products plays a critical role. It helps assess the accuracy and reliability of forecasted weather conditions by comparing them to the actual observations at specific locations. In this section, we build a global weather station sub-dataset utilizing the integrated surface dataset distributed by the National Centers for Environmental Information (NCEI), which contains surface wind speed and temperature of 18150 worldwide stations, covering 2022. We compare the predictive capabilities of three deterministic models regarding site-specific forecasting. To ensure consistency with IFS-HRES, a renowned deterministic model, our assessment focused solely on two time points, namely 00:00:00 and 12:00:00 (UTC time) of each day.

Figure~\ref{fig:observatio_tmp} presents an overall view of the station comparison result of three models from July to December 2022, with lead time increasing from 2 to 10 days. Compared with the IFS-HRES and Pangu-weather on the global station temperature forecast, FengWu-GHR yields more accurate predictions. As the lead time increases, the improvement becomes more obvious. The most noteworthy is that the temperature prediction error in December is reduced by about 0.9 degrees Celsius compared with IFS-HRES. Improvements in station forecast make FengWu-GHR potentially well-suited for various operational scenarios, including but not limited to wind energy assessment, runoff prediction, pollution control, and disaster prevention.    

\subsection{Extreme heat wave and winter storm}

\begin{figure}[htbp]
  \centering      \includegraphics[width=0.999\textwidth]{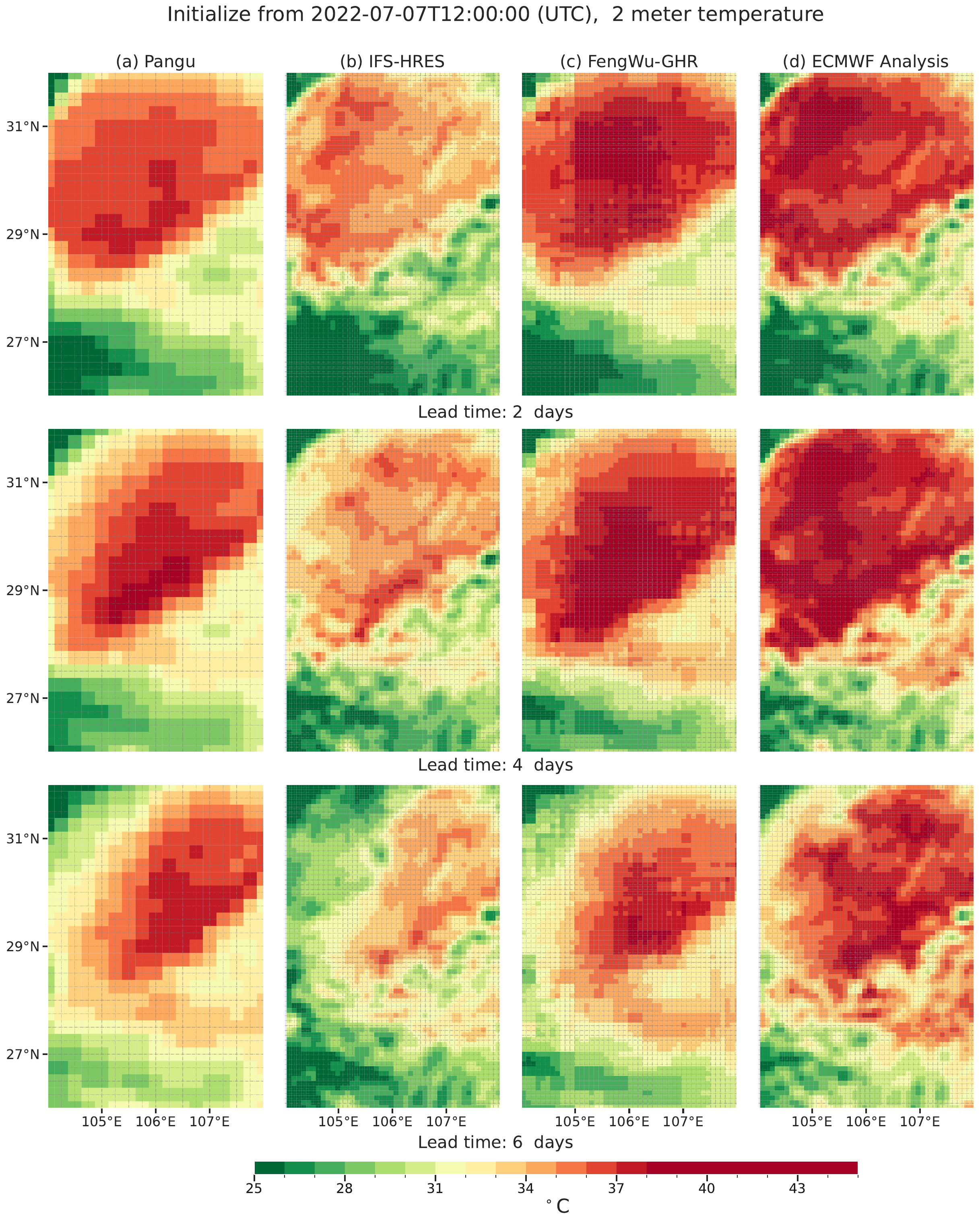}
    \caption{Surface temperature forecast of (a) Pangu-weather with $0.25^{\circ}$ spatial resolution, (b) IFS-HRES, (c) FengWu-GHR. They are initialized at 2022-07-07T12:00:00 UTC. (d) Operational analysis: it could be regarded as the ground truth assimilated from the observation data at the valid time.}
    \label{fig:resolution_compare}
\end{figure}

\begin{figure}[htbp]
  \centering      \includegraphics[width=0.84\textwidth]{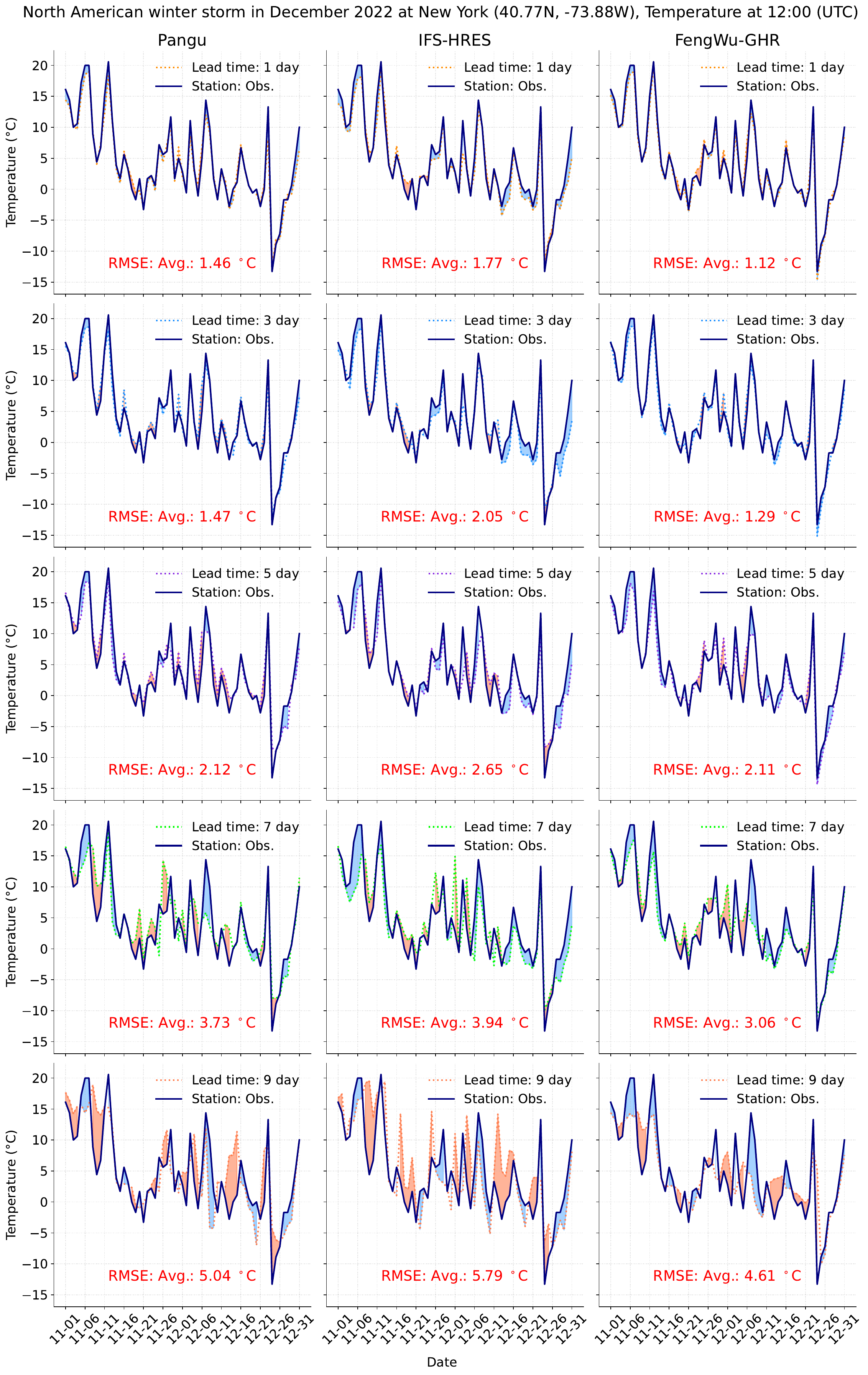}
    \caption{A severe winter storm event prediction in Dec. 2022. The temperature prediction at 2-meter at the New York surface weather station as a function of the forecast lead time in the Pangu-weather (a), IFS-HRES (b), and FengWu-GHR (c).}
    \label{fig:winter_storm}
\end{figure}

\textbf{Heat Wave}. 
During the summer of 2022, unprecedented and significant temperature anomalies were observed in Southern Europe, Western Asia, and Eastern China, exceeding the interannual standard deviation by more than twofold. These regions experienced prolonged heatwaves, especially in Chongqing, China, where daily maximum temperatures surpassed 40°C for over twenty consecutive days, repeatedly breaking historical records of extreme temperatures. However, many state-of-the-art global NWP models cannot accurately predict such extreme events (Figure~\ref{fig:resolution_compare} b), even at higher resolutions.

Figure~\ref{fig:resolution_compare} displays the heat map of t2m forecasts centered on Chongqing city ($29.5^{\circ} N, 106.5^{\circ} E$) by using different weather prediction models. Pangu-weather operates on a horizontal resolution of $0.25^{\circ}$, while the results from other models are derived from high-resolution products with a grid resolution of $0.09^{\circ} \times 0.09^{\circ}$. It is interesting that both the FengWu-GHR and Pangu, the ML-based models, can accurately predict this extreme event with a significant lead time, particularly in terms of intensity. However, FengWu-GHR accurately captures the spatial structure characteristics, center position changes, and even the evolution process of the heatwave event, which is crucial for understanding and mitigating the impact of heatwaves. Such detailed forecast products are not currently available with other low-resolution forecasts and dynamical models. This highlights the value of FengWu-GHR in providing more precise and informative predictions for extreme weather events.

\noindent \textbf{Winter Storm}. 
In addition to heatwaves, winter storms are an essential part of extreme events that have a widespread impact, occur frequently, and cause significant losses. We investigate the North American winter storm in December 2022, which affected parts of the United States and Canada from December 21st to 26th. The affected regions experienced severe weather conditions, including heavy snow, strong winds, and record-breaking low temperatures, resulting in at least 106 fatalities.

In Figure~\ref{fig:winter_storm}, FengWu-GHR is compared against two operational models IFS-HRES and Pangu, which depicts the change in temperature observation and prediction from November 1st to December 31st in New York, with comparison results shown at the lead time of 1,3,5, and 9 days. It is found that all methods can capture the abrupt temperature drop on December 23 and 24 within three lead days. However, FengWu-GHR stands out by accurately predicting the minimum temperature of this winter storm as early as nine days in advance, surpassing the performance of other models.  
Specifically, the global mean temperature trajectory error of FengWu-GHR is reduced by $18.0\%$ and $22.3\%$, respectively, compared with Pangu and IFS-HRES at one week lead time. Those results evidence that FengWu-GHR could be a more accurate operational model for such a heavy winter storm.

\section{Conclusions}\label{sec5}
This study presents the development of FengWu-GHR, the first machine learning-based global numerical weather prediction model at the highest resolution. 
It successfully tackles several challenges through advanced model design and innovative training strategies, such as the lack of high-resolution training data and overly smooth prediction problems. Notably, the core technology of learning the HR model from a large pretrained LR model is a plug-and-play approach that can be applied to most of the ML-based NWP models.

FengWu-GHR can reach or even surpass the prediction capabilities of the current state-of-the-art IFS-HRES using only 5 years of high-resolution data (2016-2021). Compared with other models represented by Pangu-weather, FengWu-GHR shows significant improvements, including more accurate and detailed descriptions of weather events, lower errors at global weather stations, and reduced bias drift. Furthermore, under the high-resolution prediction results, we novelly found that the model's ability to capture extreme events in long-lead time forecasts was enhanced. It highlights the importance of improving model resolution as one of the key directions for the development of ML-based weather forecast models.

Moreover, it is noted that the current version of FengWu-GHR is considered preliminary. Future versions will incorporate additional variables, including precipitation, which is of significant interest, as well as variables related to estimations of solar and wind power. These additions will provide a more comprehensive set of information for users.

\section*{Acknowledgements}
The dataset used in this paper is supported by the European Centre for Medium-Range Weather Forecasts. We use the information at our own sole risk and liability. The ECMWF has no liability in respect of this document, which is merely representing the author's view.

We would like to express our gratitude to the Research Support, IT, and Infrastructure team at the Shanghai AI Laboratory for their valuable assistance in providing computation resources and network support throughout this research project. We also acknowledge the support of F Ling and J-J Luo, who are funded by the National Key Research and Development Program of China (No. 2020YFA0608000). We thank Xinyu Wang and Wanghan Xu for their participation in discussions.

We would like to extend our special appreciation to the team members, namely Prof. Yu Qiao, Qihong Liao, Jiamin Ge, Jing Zou, Jingwen Li, and Xingpu Li, for their support. 

\bibliographystyle{unsrtnat}
\bibliography{sn-bibliography}

\clearpage

\end{document}